\documentclass{article}
\usepackage{spconf,amsmath,epsfig}
\usepackage[table,xcdraw]{xcolor}
\usepackage{lineno}
\usepackage{soul}
\usepackage{xcolor}
\usepackage[hidelinks]{hyperref}
\usepackage{fancyhdr}
\usepackage{url}

\fancypagestyle{IEEE}{
\setlength{\headheight}{50pt} % space to title
\addtolength{\topmargin}{-55pt} % raise IEEE notice to top edge
\fancyhf{}
\fancyhead[C]{
Copyright notice: \copyright 2022 IEEE. Personal use of this material is permitted. Permission from IEEE must be obtained for all other uses, in any current or future media, including reprinting/republishing this material for advertising or promotional purposes, creating new collective works, for resale or redistribution to servers or lists, or reuse of any copyrighted component of this work in other works. This material is referenced by DOI: \href{https://doi.org/10.1109/IGARSS46834.2022.9883198}{10.1109/IGARSS46834.2022.9883198} 
}
}

\title{RapidAI4EO: Mono- and Multi-temporal Deep Learning models for\\ Updating the CORINE Land Cover Product}

\name{
\parbox[c]{\textwidth}{
\centering
P. Bhugra, B. Bischke, C. Werner, R. Syrnicki, C. Packbier, P. Helber, \\
C. Senaras, A. S. Rana, T. Davis, W. De Keersmaecker, D. Zanaga,\\
A. Wania, R. Van De Kerchove, G. Marchisio
}
}

\address{
(priyash.bhugra, benjamin.bischke, christoph.werner, \\ carolin.packbier, robert.syrnicki, patrick.helber)@vision-impulse.com\\ 
(caglar.senaras, akhil.singh.rana, tim.davis, annett.wania)@planet.com\\
(wanda.dekeersmaecker, daniele.zanaga, ruben.vandekerchove)@vito.be\\
giovanni@planet.com
}

\begin{document}
\thispagestyle{IEEE}
%FIXME!!
\maketitle
\begin{abstract}
In the remote sensing community, Land Use Land Cover (LULC) classification with satellite imagery is a main focus of current research activities. Accurate and appropriate LULC classification, however, continues to be a challenging task. In this paper, we evaluate the performance of multi-temporal (monthly time series) compared to mono-temporal (single time step) satellite images for multi-label classification using supervised learning on the RapidAI4EO dataset.
As a first step, we trained our CNN model on images at a single time step for multi-label classification, i.e.\ mono-temporal. We incorporated time-series images using a LSTM model to assess whether or not multi-temporal signals from satellites improves CLC classification. The results demonstrate an improvement of approximately 0.89\% in classifying satellite imagery on 15 classes using a multi-temporal approach on monthly time series images compared to the mono-temporal approach. Using features from multi-temporal or mono-temporal images, this work is a step towards an efficient change detection and land monitoring approach.    
\end{abstract}

\section{Introduction}
\label{sec:intro}

Remote sensing plays an important role in the observation and study of the earth’s surface by providing mono-temporal and multi-temporal information on land use/land cover (e.g., artificial area, forest, bare ground, agriculture, etc.). Land use is determined by how humans use it for recreation, wildlife habitats, urban development and agriculture. Whereas, land cover is the extent to which the Earth's surface is covered by forests, wetlands, urban, and agriculture areas~\cite{Prakasam2010LandUA}. Land Use and Land Cover (LULC) is a method of categorizing and classifying human and natural activities in the landscape over a specific period. In the past decades, many Earth observing satellites like Sentinel-2 by European Space Agency, Landsat 7 and 8 by U.S. Geological Survey, etc have come into existence. The data from such satellites were used by researchers to analyze and track land use and land changes over time in response to human factors such as climate change, urbanization, deforestation, wildfires, etc. RapidAI4EO~\cite{RAPIDAI4EO}, a new satellite imagery corpus, aims to provide better monitoring of Land Use (LU), Land Cover (LC), and LULC change at a much higher level of detail and temporal frequency than is currently possible~\cite{RAPIDAI4EO}. The objective of this paper is to explore the advantages of using multi-temporal (monthly) over mono-temporal (single time-step) satellite images. To achieve this, we trained our models with different land cover ontologies. Level-1: 5 classes, Level-1.5: 7 classes, Level-2: 15 classes with single time-step as well as high cadence i.e., monthly imagery from RapidAI4EO dataset.

\section{Related Work}
\label{sec:related}

\subsection{LULC classification}

A common application of remote sensing is creating land-use/land-cover classification maps (LULC) on satellite images. As satellite imagery becomes increasingly affordable and more precise, many government and private industries are using LULC classification maps for a wide range of applications, including land monitoring for human or natural activities, damage delineation (e.g. fire, flood forecasting), wildlife habitat preservation and planning rural and urban land use. Deep learning-based image classification models categorize an image belonging to one or more categories such as single class, multi-class and multi-label classification. The main focus of this research is to learn the multi-label patch-based distribution which can classify satellite images (e.g. 40\% agriculture, 30\% forest, 30\% water bodies) using images at single timestep as well as images with monthly cadence. To achieve this, we used mono-temporal and multi-temporal approaches using a supervised classifier to extract features from multi-spectral satellite images. We contrast on the state-of-the-art Convolutional Neural Network (CNN)~\cite{CNN} architecture such as ResNet-50~\cite{Resnet-50} to train a patch-based multi-label classifier that recognizes multiple classes from a single imagery.

\subsection{Mono-temporal approach}
\begin{figure}[ht!]
    \centering
     \includegraphics[width=85mm]{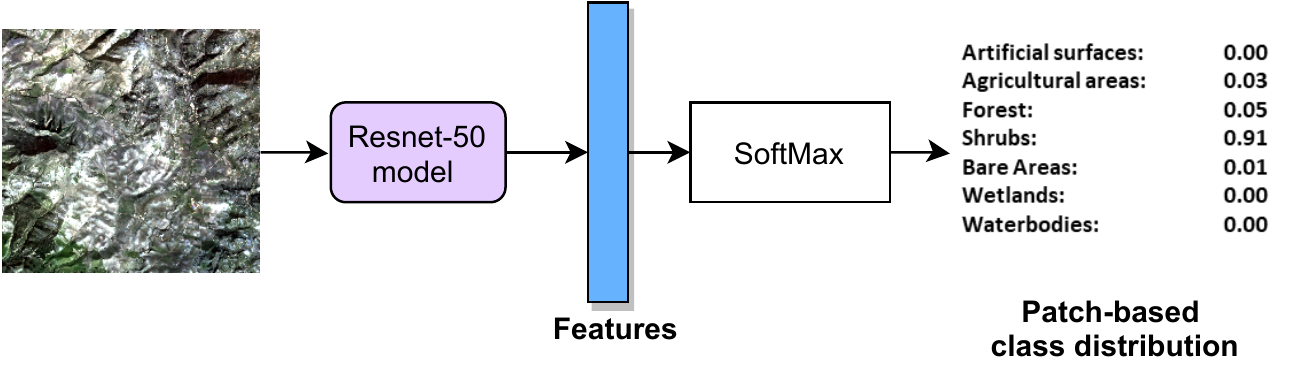}
     \caption{\textbf{Mono-temporal architecture}: It passes a single time step satellite image as input to the ResNet-50 model and outputs a multi-label patch-based distribution.}
     \label{fig:monotemporal}
\end{figure}

The mono-temporal approach uses images at a single time step (the monthly mosaic images) from the RapidAI4EO dataset, consisting of Planet Fusion (PF) and Sentinel-2 (S2) imagery as an input to the ResNet-50 model. Figure~\ref{fig:monotemporal} represents the architecture, in which features are extracted from a single time step image using ResNet-50, and then SoftMax is applied for multi-label distribution.

\subsection{Multi-temporal approach}
\begin{figure}[ht!]
    % \centering
     \includegraphics[width=78mm]{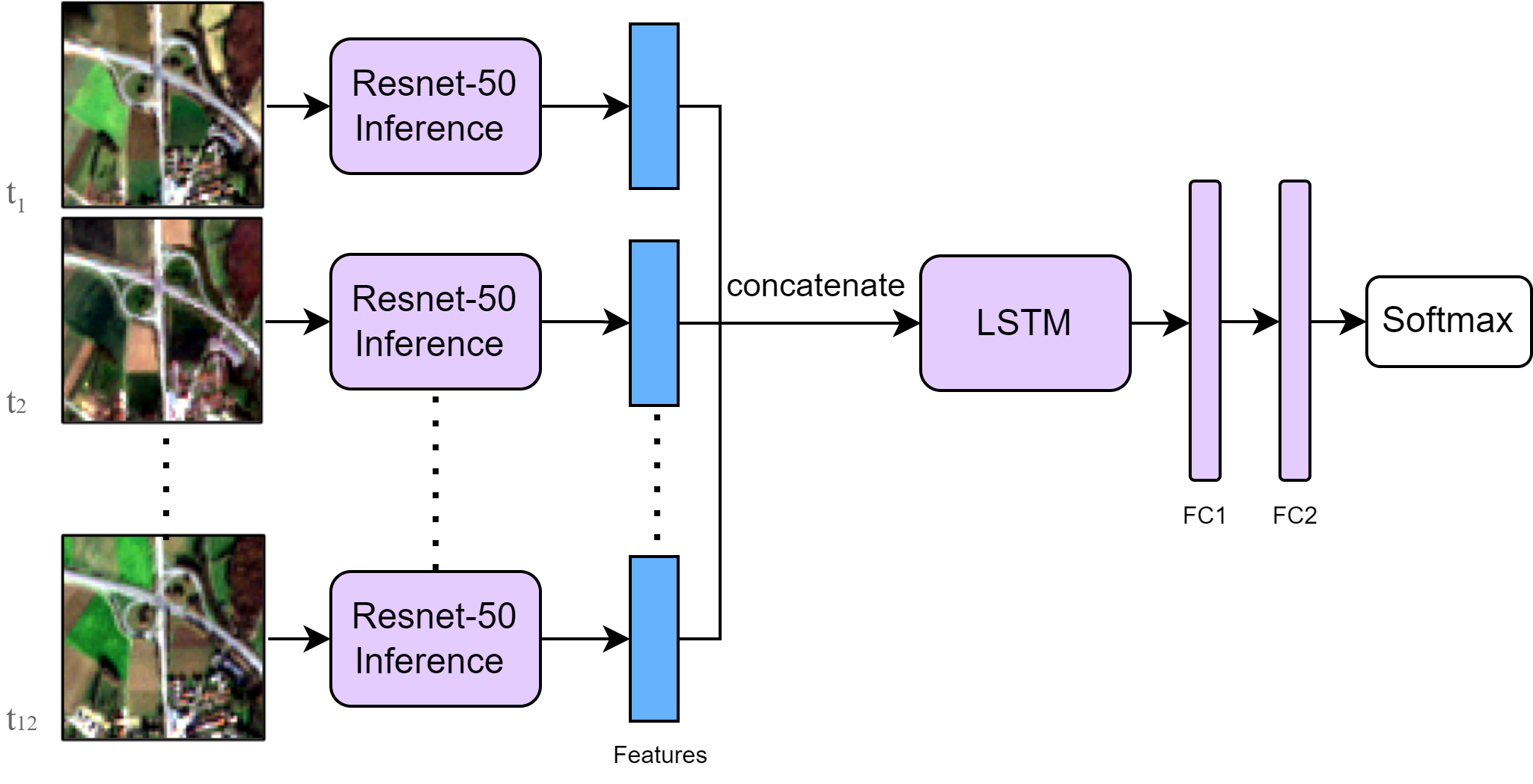}
     \caption{ \textbf{Multi-temporal architecture}: It takes monthly images over the year as input to the ResNet-50 model, later the features from the model is taken as an input to LSTM with 2 fully connected later which outputs a multi-label patch-based distribution.}
     \label{fig:multi-temporal}
\end{figure}

As a way of exploring the implications of multi-temporal data, we passed the existing ResNet-50 features trained on mono-temporal images into an LSTM~\cite{LSTM}. Figure~\ref{fig:multi-temporal} illustrates the architecture of the multi-temporal approach, in which the features are extracted from multi-temporal images using ResNet-50, concatenated and then passed to an LSTM. Later, the output is passed through two fully connected layers and then post-processed by applying softmax for patch-based multi-label distribution. The multi-temporal satellite images with monthly cadence provided us with a chance to investigate if it could further improve the multi-label LULC classification. \\

\subsection{Training configuration}
As part of the ResNet-50 training process, we used RapidAI4EO dataset, which covers 500,000 patches across the European Economic Area (EEA39), with a 70:20:10\% split for training, validation, and testing respectively. These 500,000 locations comprise 1400 tiles where each tile covers an area of 8000 x 8000 meters. We further divide this 8000 x 8000 m area into 1600 small patches, on which we ran our ResNet-50 model. The ResNet-50 model is pretrained on ImageNet. Four-channel (RGB-N) images were taken as input for our model. The images were normalized using the mean and standard deviation calculated over all training images. The model is trained for 20 epochs with a learning rate of 0.0001. After every epoch, the learning rate is reduced by a factor of 0.1. We experimented with different loss functions such as Focal Loss, BCE loss and KL Divergence loss for training our model concluding KL Divergence loss to be the best. To incorporate the multi-temporal approach, we trained our LSTM model for 20 epochs with a learning rate of 0.00001 using KL Divergence loss. 

\section{Experiments}
In this section, we explore the new dataset for LULC classification. Furthermore, we ran experiments to show the improvement of multi-temporal times series data over mono-temporal single time step image.

\subsection{Dataset}
\label{sec:format}
RapidAI4EO~\cite{RAPIDAI4EO} will establish the basis for the next-generation Copernicus Land Monitoring Service (CLMS) under the sponsorship of the European Union's Horizon 2020 program. The idea of such a large corpus comes from patch-based LULC EuroSAT~\cite{eurosat} and BigEarthNet~\cite{bigearthnet} corpora. RapidAI4EO aims at creating a dense spatio-temporal training corpus by combining S2 with high cadence, very high resolution and harmonized multi-spectral PF imagery at 500,000 patch locations spread over EEA39.

\begin{figure}[h!]
    \centering
     \includegraphics[width=70mm,height=40mm]{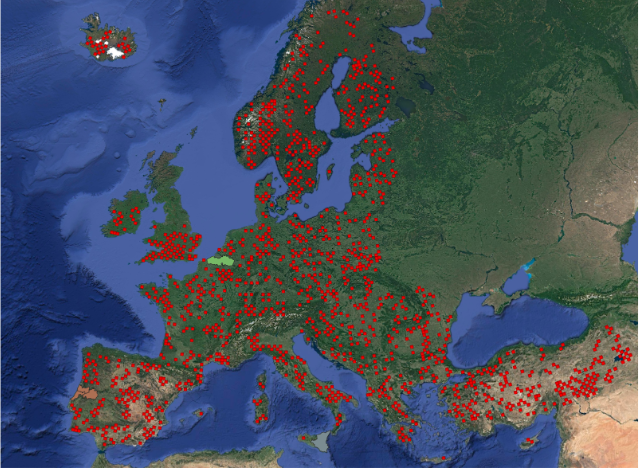}
     \caption{ The map shows the sampling of tiles spread over EEA39. Red points represent the overall 500,000 selected locations.}
     \label{fig:tiles}
\end{figure}
RapidAI4EO dataset consists of PF (3-meter spatial resolution) and S2 (10-meter spatial resolution) images for selected European locations. These selected locations are sampled across EEA39 to ensure that each country covered by Corine Land Cover 2018 (CLC 2018) is adequately represented, that there is a minimum number of CLC 2018 samples per class, and that the CLC 2018 class distribution is preserved. For our approach, we sampled 500,000 locations, which can be seen in Figure~\ref{fig:tiles}, the red points correspond to the overall 1400 tiles used for the production of 500,000 locations sampled across EEA39.

\subsection{Mono-temporal experiments}
\label{sec:monotemporal}
To begin, we trained our ResNet-50 model on RapidAI4EO dataset, i.e. 500,000 locations across EEA39 with the different land cover ontologies, Level-1 (5 classes), Level-1.5 (7 classes) and Level-2 (15 classes). This CLC nomenclature is organized into multiple levels of hierarchical classification, with 44 classes at the most detailed level and 5 classes at the least, is defined here~\footnote{\url{https://land.copernicus.eu/eagle/files/eagle-related-projects/pt_clc-conversion-to-fao-lccs3_dec2010}}. As the forest and semi-natural area class from Level-1 ontology contains visually very distinctive subclasses, it can be further separated into forests, shrubs, and bare areas, leading to 7 classes. The aggregated class labels are also mapped as a probability distribution like [0.4,0.5,0,0.1,0] rather than a binary one hot encoding [1,1,0,0,0]. Table~\ref{tab:clc_nom_resnet_50} shows that our classification model achieves an F1-score (micro) on test set of 60.78\% for 15 classes and up to 73.88\% for 7 classes. When the land cover classes are aggregated into 5 classes, the F1-score (micro) rises to 81.24\%.

\begin{table}[ht!]
\centering
\scalebox{0.70}{
\begin{tabular}{|c|c|c|c|c|c|}
\hline
\rowcolor[HTML]{A4C2F4} 
\multicolumn{1}{|l|}{\cellcolor[HTML]{A4C2F4}\textbf{Method}} &
  \multicolumn{1}{l|}{\cellcolor[HTML]{A4C2F4}\textbf{Bands}} &
  \multicolumn{1}{l|}{\cellcolor[HTML]{A4C2F4}\textbf{Dataset}} &
  %\multicolumn{1}{l|}{\cellcolor[HTML]{A4C2F4}\textbf{Cadence}}
  \multicolumn{1}{l|}{\cellcolor[HTML]{A4C2F4}\textbf{Classes}} &
  \multicolumn{1}{l|}{\cellcolor[HTML]{A4C2F4}\textbf{F1-score (Micro) in \%}} \\ \hline
Mono-temporal & RGB-N & RapidAI4EO  & 15 & 60.78          \\ \hline
Mono-temporal & RGB-N & RapidAI4EO & 7 &  73.88     \\ \hline
Mono-temporal & RGB-N & RapidAI4EO  & 5  & 81.24   \\ \hline
\end{tabular}}
\caption{The table illustrates the importance of class aggregation on RapidAI4EO dataset. As evident, after aggregating classes, the F1-score of our model increased from 60.78\% to 81.24\%.}
\label{tab:clc_nom_resnet_50}
\end{table}

\subsection{Multi-temporal experiments}

To implement the multi-temporal approach, we experiment with monthly images  i.e.\ 15th of every month, on 3 different land cover ontologies. As observed in Table~\ref{tab:weekly_monthly}, the multi-temporal model achieves an F1-score of 61.67\% for 15 classes and up to 74.67\% for 7 classes. The overall performance increases to 81.69\% when the land cover classes are aggregated into 5 classes.

\begin{table}[ht!]
\centering
\scalebox{0.62}{
\begin{tabular}{|c|c|c|c|c|c|}
\hline
\rowcolor[HTML]{A4C2F4} 
\multicolumn{1}{|l|}{\cellcolor[HTML]{A4C2F4}\textbf{Method}} &
  \multicolumn{1}{l|}{\cellcolor[HTML]{A4C2F4}\textbf{Bands}} &
  \multicolumn{1}{l|}{\cellcolor[HTML]{A4C2F4}\textbf{Dataset}} &
  \multicolumn{1}{l|}{\cellcolor[HTML]{A4C2F4}\textbf{Cadence}} &
  \multicolumn{1}{l|}{\cellcolor[HTML]{A4C2F4}\textbf{Classes}} &
  \multicolumn{1}{l|}{\cellcolor[HTML]{A4C2F4}\textbf{F1-score (Micro) in \% }} \\ \hline
Multi-temporal & RGB-N & RapidAI4EO & Monthly & 15 & 61.67            \\ \hline
Multi-temporal & RGB-N & RapidAI4EO & Monthly & 7 & 74.67             \\ \hline
Multi-temporal & RGB-N & RapidAI4EO & Monthly & 5  & 81.69   \\ \hline
\end{tabular}}
\caption{The table shows the detailed experiment on monthly time-series cadence using the multi-temporal approach.}
\label{tab:weekly_monthly}
\end{table}

\subsection{Comparison of mono-temporal and multi-temporal}

\begin{table}[ht!]
\scalebox{0.85}{
\begin{tabular}{lll}
\hline
\rowcolor[HTML]{A4C2F4} 
\multicolumn{1}{|l|}{\cellcolor[HTML]{A4C2F4}\textbf{Class Name}}                              & \multicolumn{1}{l|}{\cellcolor[HTML]{A4C2F4}\textbf{\begin{tabular}[c]{@{}l@{}}Mono-temporal \\ (F1-score in \%)\end{tabular}}} & \multicolumn{1}{l|}{\cellcolor[HTML]{A4C2F4}\textbf{\begin{tabular}[c]{@{}l@{}}Multi-temporal Monthly \\ (F1-score in \%)\end{tabular}}} \\ \hline
\multicolumn{1}{|l|}{Urban Fabric}                                                             & \multicolumn{1}{c|}{\textbf{48.96}}                                                                                             & \multicolumn{1}{c|}{47.12}                                                                                                               \\ \hline
\multicolumn{1}{|l|}{Industrial Units}                                                         & \multicolumn{1}{c|}{\textbf{35.46}}                                                                                             & \multicolumn{1}{c|}{32.35}                                                                                                               \\ \hline
\multicolumn{1}{|l|}{Arable Land}                                                              & \multicolumn{1}{c|}{72.84}                                                                                                      & \multicolumn{1}{c|}{\textbf{73.96}}                                                                                                      \\ \hline
\multicolumn{1}{|l|}{Permanent crops}                                                          & \multicolumn{1}{c|}{51.68}                                                                                                      & \multicolumn{1}{c|}{\textbf{55.94}}                                                                                                      \\ \hline
\multicolumn{1}{|l|}{Pastures}                                                                 & \multicolumn{1}{c|}{29.20}                                                                                                      & \multicolumn{1}{c|}{\textbf{32.19}}                                                                                                      \\ \hline
\multicolumn{1}{|l|}{Forests}                                                                  & \multicolumn{1}{c|}{\textbf{78.11}}                                                                                             & \multicolumn{1}{c|}{77.70}                                                                                                               \\ \hline
\multicolumn{1}{|l|}{\begin{tabular}[c]{@{}l@{}}Scrubs, herbaceous\\ vegetation\end{tabular}}  & \multicolumn{1}{c|}{50.29}                                                                                                      & \multicolumn{1}{c|}{\textbf{53.01}}                                                                                                      \\ \hline
\multicolumn{1}{|l|}{\begin{tabular}[c]{@{}l@{}}Open space with \\ no vegetation\end{tabular}} & \multicolumn{1}{c|}{65.05}                                                                                                      & \multicolumn{1}{c|}{\textbf{67.87}}                                                                                                      \\ \hline
\multicolumn{1}{|l|}{Inland wetlands}                                                          & \multicolumn{1}{c|}{38.79}                                                                                                      & \multicolumn{1}{c|}{\textbf{42.63}}                                                                                                      \\ \hline
\multicolumn{1}{|l|}{Inland waters}                                                            & \multicolumn{1}{c|}{\textbf{69.85}}                                                                                             & \multicolumn{1}{c|}{69.82}                                                                                                               \\ \hline
                                                                                               & \multicolumn{1}{l}{}                                                                                                            & \multicolumn{1}{l}{}                                                                                                                    
\end{tabular}}
\caption{Class-wise comparison of mono and multi-temporal approaches based on F1-score.}
\label{tab:mon_week}
\end{table}
To demonstrate the advantage of the multi-temporal approach over the mono-temporal approach, we compared the mono-temporal approach on single time step images with the multi-temporal approach on monthly images. As can be observed from Tables~\ref{tab:clc_nom_resnet_50} and~\ref{tab:weekly_monthly}, multi-temporal classification of monthly time-series images outperformed those based solely on mono-temporal classification. The multi-temporal model on monthly images achieves the overall improvement in F1-score of 0.89\% for 15 classes, 0.79\% for 7 classes and 0.45\% for 5 classes when compared with mono-temporal model. As evidence from Table~\ref{tab:mon_week}, the mono-temporal approach shows minimal advantage over multi-temporal for classes such as urban, industrial forest areas and inland waters.

\section{Model Inference and Applications}
\begin{figure}[t!]
    \centering
     \includegraphics[width=80mm]{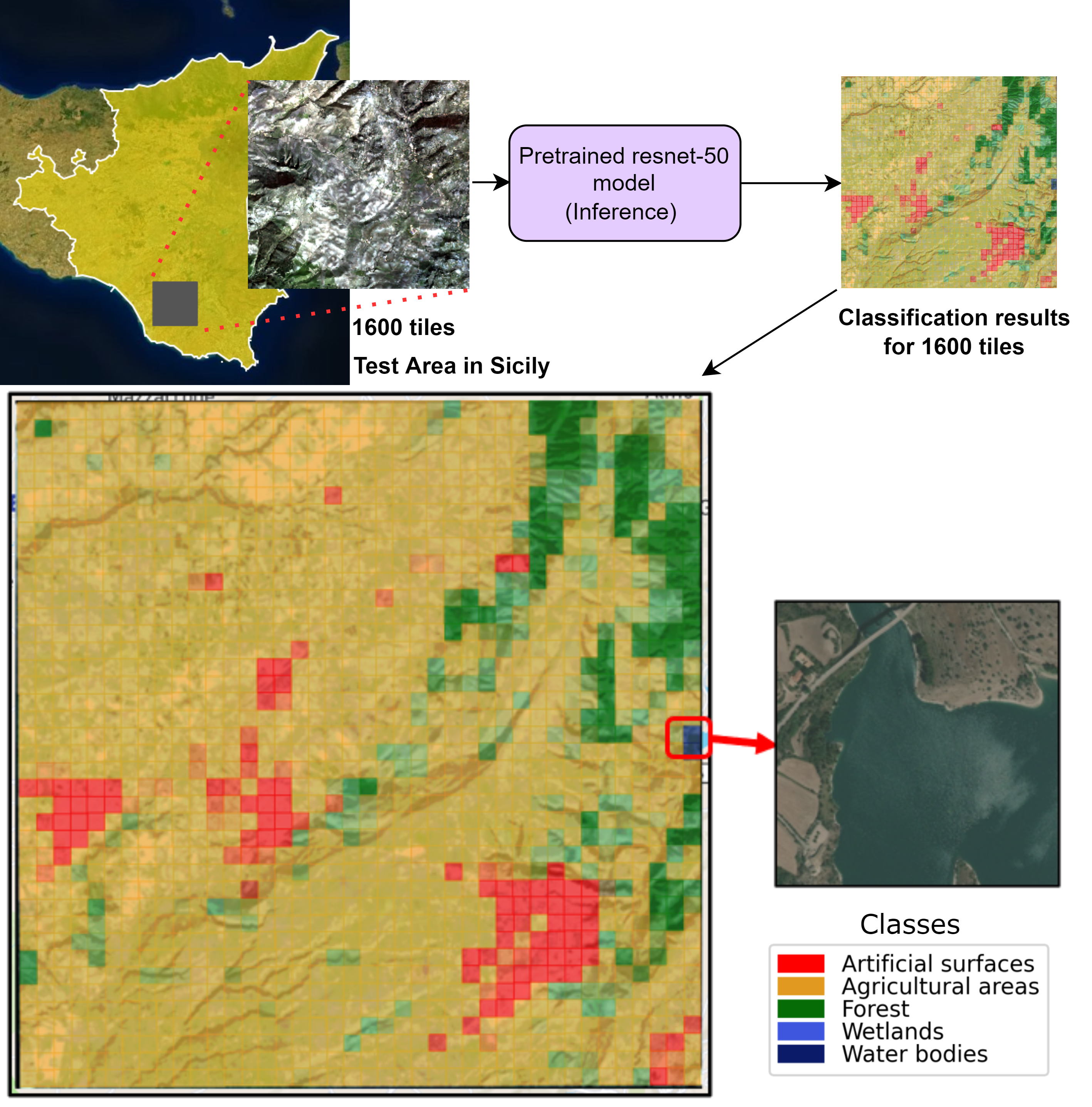}
     \caption{Classification map of test area in Sicily with a highlighted tile for example.}
     \label{fig:test_tiles}
\end{figure}

\begin{figure}[t!]
    \centering
     \includegraphics[width=80mm]{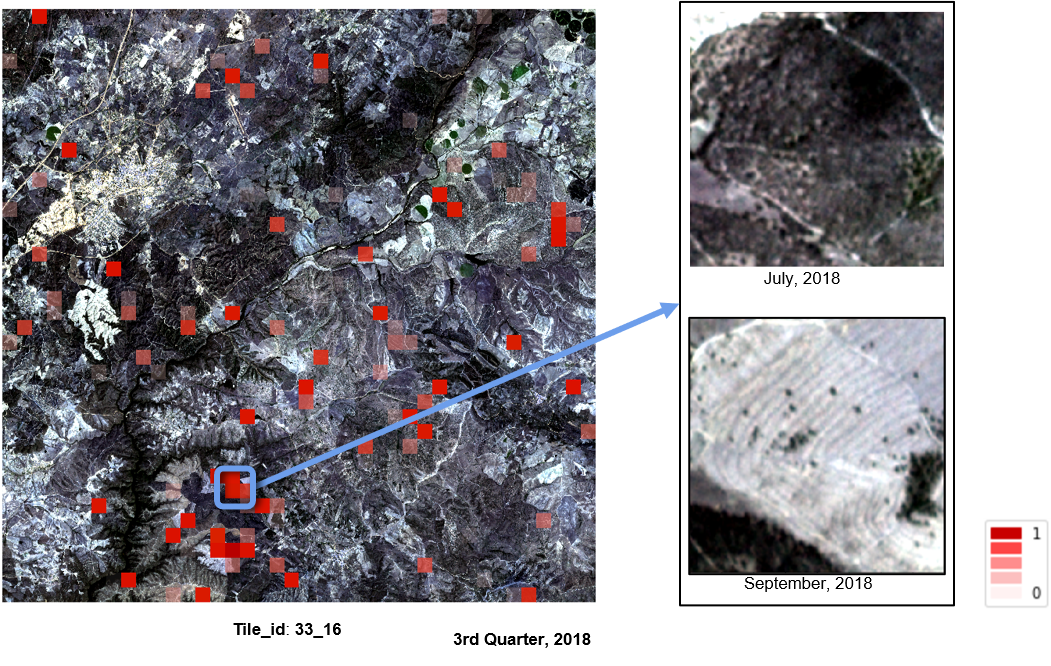}
     \caption{Change detection map of test area in Portugal with a highlighted example.}
     \label{fig:cd}
\end{figure}

To evaluate the performance of our model we selected a test area that has not been used in the model training. We generated LULC maps on the test area in Sicily as shown in Figure~\ref{fig:test_tiles}. As can be observed, our model is able to classify agriculture, forest, water bodies and artificial areas precisely. These LULC maps can be used for land monitoring, planning urban land use, wildlife habitat preservation and change detection. We also showed an example of how we can use our model for the change detection part in Figure~\ref{fig:cd}. 
\label{sec:typestyle}
    
\section{CONCLUSION}
\label{sec:majhead}
In this paper, we compared the LULC classification using mono-temporal i.e.\ single time step images and multi-temporal images on monthly cadence. For the mono-temporal approach, we train a ResNet-50 model on the RapidAI4EO dataset for multi-label classification. To incorporate multi-temporal images, we used the mono-temporal model for inference on monthly images and passed it through a LSTM. We can conclude from the evaluation that the multi-temporal approach outperforms the mono-temporal approach on F1-score (micro) by approximately 0.89\%. Nevertheless, it is obvious from Table~\ref{tab:mon_week} that the mono-temporal approach is suitable for only a few classes while the multi-temporal approach is suitable for most of the classes. These LULC maps can be leveraged for multiple real-world Earth observation applications. A few application areas include detection of changes in land use and land cover, delineation of damaged areas, and urban planning.

\section{ACKNOWLEDGEMENTS}
\label{sec:print}

This project has received funding from the European Union’s Horizon 2020 research and innovation programme under grant agreement No 101004356.

\label{sec:page}

\bibliographystyle{IEEEbib}
\bibliography{refs}

\end{document}